\documentclass[a4paper,12pt,twoside]{report}
\usepackage[left=2.5cm,right=2.5cm,top=2cm,bottom=3cm]{geometry}
\usepackage{natbib}
\usepackage{amsmath}
\usepackage{graphicx,changepage}
\usepackage{graphics}
\usepackage{amssymb}
\usepackage{caption}
\usepackage{physics}
\usepackage{subcaption}
\usepackage{url}
\usepackage{multirow,multicol}

\usepackage{graphicx}
\usepackage{verbatim}
\usepackage{latexsym}
\usepackage{mathchars}
\usepackage{setspace}
\usepackage[toc]{appendix}

\input{blocked.sty}
\input{uhead.sty}
\input{boxit.sty}
\input{thesis.sty}

\newcommand{\NN}{{\sf I\kern-0.14emN}}   
\newcommand{\ZZ}{{\sf Z\kern-0.45emZ}}   
\newcommand{\QQQ}{{\sf C\kern-0.48emQ}}   
\newcommand{\RR}{{\sf I\kern-0.14emR}}   

\newcommand{\normallinespacing}{\renewcommand{\baselinestretch}{1.5} \normalsize}

\newcommand{\syncc}{~\stackrel{\textstyle \rhd\kern-0.57em\lhd}{\scriptstyle L}~}

\setcitestyle{square}

\DeclareMathOperator*{\argminA}{arg\,min} 
\DeclareMathOperator*{\argmaxA}{arg\,max} 

\begin{document}

\title{Learning Domain Specific Language Models
for Automatic Speech Recognition through
Machine Translation
}

\author{Saurav Jha}
\submitdate{Jun 2021}
\degree{MSc Advanced Systems Dependability}
\studentid{32029567}
\supervisor{Imran Sheikh and Emmanuel Vincent}

\normallinespacing
\maketitle


\preface
\addcontentsline{toc}{chapter}{Abstract}

\begin{abstract}

Automatic Speech Recognition (ASR) systems have been gaining popularity in the recent years for their widespread usage in smart phones and speakers. Building ASR systems for task-specific scenarios is subject to the availability of utterances that adhere to the style of the task as well as the language in question. In our work, we target such a scenario wherein task-specific text data is available in a language that is different from the target language in which an ASR Language Model (LM) is expected. We use Neural Machine Translation (NMT) as an intermediate step to first obtain translations of the task-specific text data. We then train LMs on the 1-best and N-best translations and study ways to improve on such a baseline LM. We develop a procedure to derive word confusion networks from NMT beam search graphs and evaluate LMs trained on these confusion networks. With experiments on the WMT20 chat translation task dataset, we demonstrate that NMT confusion networks can help to reduce the perplexity of both n-gram and recurrent neural network LMs compared to those trained only on N-best translations.
\end{abstract}


\body
\chapter{Introduction}

Voice interfaces in the form of personal assistants and spoken dialogue systems have gained ubiquitous usage in the last decade. This has been possible due to the advances brought by machine learning and deep learning methods in Automated Speech Recognition (ASR), Text-to-Speech  (TTS) synthesis and Natural Language Understanding (NLU) systems. At the same time, lack of sufficient training data, for each of these components, has limited the availability of voice interfaces across several languages. Moreover, the existing interfaces are also restricted to specific domains or tasks in the given language. The work carried out in this internship lies along the research direction of enabling domain-specific or task-specific ASR in different languages.

An ASR system uses an Acoustic Model (AM) and a Language Model (LM) to perform a speech-to-text conversion. An AM trained on a generic corpus of speech recordings and corresponding text transcriptions can be reused across multiple applications. But an accurate speech transcription for a specific application requires that the LM is well trained or adapted to the target domain, for instance, medical, travel, shopping, etc. Lack of domain specific text corpora limits the development of ASR LMs and hence the deployment of accurate  and practical domain specific ASR and voice assistants. This work tries to reuse domain specific corpora from one language to build LMs for other language(s).

\section{Dependability for ASR}  For a software system to have dependable attributes, it must demonstrate availability, reliability, and confidentiality. The following points brief how learning domain specific LMs help accomplish these benchmarks for ASR systems:

\begin{enumerate}
    \item[a.] Availability: The LMs for domain specific ASR systems are usually trained on large amount of generic corpora. Switching to domain specific LMs can instead help covering a wider range of task-related queries thus increasing the readiness for service.
    
    \item[b.] Reliability: Domain-specific LMs can aid to understanding and generating relevant responses for a given task when compared to generic LMs. Such ASR systems can in turn deliver stable services for longer time than their general purpose counterparts.
    
    \item[c.] Confidentiality: A major privacy concern for ASR  remains \textit{function creep} wherein personal information collected for one purpose is extended to another. Further, the speech cues to be preserved might vary with the domain in question. Learning domain-specific LMs for ASR can aid in preserving relevant information while providing potential performance boost in the private domain \citep{kerrigan-etal-2020-differentially}.
    
\end{enumerate}

\section{Contributions}

We explore the usage of Machine Translation (MT) to translate domain specific text data in a source language (German dialogues in our case) to their domain specific counterparts in the target language (English). Prior work on training LMs on machine translated text has conventionally exploited single-best output of MT \citep{lembersky2012language}. However, the main challenge is that generic MT systems can result in less accurate transcriptions on domain specific text. Moreover, task-specific LMs are expected to model different possible formulations of spoken queries, which may not be learned only from single-best output of MT. Hence, we explore training of domain specific LMs from multiple translation alternatives obtained from MT.

Prior works in speech translation have demonstrated the superiority of using alternate ASR hypotheses in the form of lattices and word confusion networks \citep{sperber2017neural,dupont1997lattice,ney1999speech, schultz2004using}. Motivated by these works, we derive word confusion networks from N-best translations and train both n-gram and recurrent neural network LMs on them. The main contributions of this internship are.
\begin{enumerate}
    \item[A)] Training and evaluation of LMs trained on 1-best and N-best machine translation outputs.
    \item[B)] A procedure to derive word confusion networks from N-best translations of a neural MT system.
    \item[C)] Training and evaluation of methods that train LMs on word confusion networks derived from N-best translations.
\end{enumerate}

The rest of this report is organized as follows. Chapter \ref{ch:background} builds a background on ASR, MT, beam search decoding in MT and discusses related work from the literature. Chapter \ref{ch:approach} presents our problem statement, discusses our approach  to derive confusion networks from N-best translations and  to train LMs on these confusion networks. Experiments and evaluation are reported in Chapter \ref{ch:exp}, followed by the conclusion in Chapter \ref{ch:conclusion}. 
\chapter{Background}
\label{ch:background} 

\section{Automatic Speech Recognition}

\begin{figure*}[!htbp]
        \centering
        \includegraphics[width=0.8\textwidth]{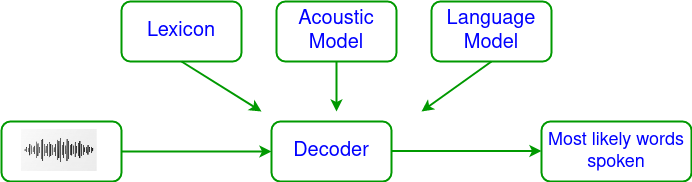}
        \caption{Components of an ASR system.}
        \label{asrsystem}
\end{figure*}

The different components of an Automatic Speech Recognition (ASR) system are depicted in Figure \ref{asrsystem}. The components involved in the system are briefly described below.
\begin{enumerate}
    \item[A)] An Acoustic Model (AM) captures the relationship between a speech signal and its constituent linguistic units such as phonemes. Most commonly used acoustic models include Hidden Markov Models (HMMs) along with Gaussian Mixture Models (GMMs) \citep{18626} and/or artificial neural networks \citep{hinton2012deep}. An AM is trained on a dataset consisting of pairs of spoken utterances and their corresponding transcriptions.
    \item[B)] Language Models (LMs) in ASR express the likeliness $P(W)$ of a word sequence to be a source sentence of the language in question. In essence, they model the probability of the occurrence of a word $w_t$ given the previous $\rho$ tokens, \textit{i.e.,}  $\log  P(w_t | w_{t-1}, ..., w_{t - \rho})$. 

        ASR LMs are typically n-gram LMs or neural LMs. 
        An n-gram LM computes $\log  P(w_t | w_{t-1}, ..., w_{t - \rho})$ based on n-gram counts (Section \ref{ngramlmcn2lm}), limiting $\rho$ = $n - 1$. Most ASR systems use 3-gram or 4-gram LMs. Neural LMs represent words by continuous vector representations and commonly use feed-forward neural network or Recurrent Neural Network (RNN) architectures to model $\log  P(w_t | w_{t-1}, ..., w_{t - \rho})$. RNNs have the capacity to model long history of word sequences, with theoretically no limits on $\rho$.
    \item[C)] A lexicon comprises a mapping of word tokens to their phonetic transcriptions. In traditional vocabulary-based ASR systems, a lexicon establishes the link between the phone predictions made by the AM and the word sequences scored using an LM.
    \item[D)]A decoder combines the aforementioned components to decode a sequence of words that best matches the input speech signal. Given a sequence of acoustic observations $O = o_1, o_2, ..., o_t$ representing the temporally consecutive slices of the speech signal, the decoder hypothesizes the optimal sequence of words $W = w_1, w_2, ..., w_n$ composing the input signal as:
    \begin{equation}
        \hat{W} = \argmaxA_W P(W|O) = \argmaxA_W P(W) P(O|W)
    \end{equation}
    where $P(O|W)$ and $P(W)$ denote the \textit{observation likelihood} and the \textit{prior probability}, and are computed by the acoustic model and the language model, respectively. 
\end{enumerate}

\section{Machine Translation} 
The very first approaches to Machine Translation (MT) include rule-based systems which leverage an exhaustive set of morphological, syntactic and semantic rules for the synthesis of target language texts from source language texts \citep{forcada2011apertium}.  This was followed by the early statistical machine learning-driven corpus-based MT systems wherein a learning algorithm is supervised by a large body of parallel corpora and then employed to translate previously unseen sentences \cite{koehn2009statistical}. A yet another line of MT systems include the hybrid approaches which combine the statistical and rule-based methods. The major lines of work along this branch include word-based \citep{brown1993mathematics} and phrase-based \citep{och2003minimum} MT systems. 

Deep learning based approaches, commonly referred to as Neural Machine Translation (NMT), have become mainstream for MT in the last few years, . These approaches train an Encoder-Decoder model wherein an encoder neural network encodes the source language text into a vector representation and the decoder decodes the target language text from the encoded vector \citep{cho2014properties}. In general, an attention mechanism is applied to the hidden state vectors of the encoder neural network \citep{luong2015effective}. Until recently, RNNs  were the common choice for encoder and decoder units. Recently, the rise of Transformers-based \citep{vaswani2017attention} architectures have largely replaced RNNs in state-of-the-art neural MT models.


\section{Decoding and Beam Search}
\label{beamsearchgraphs}

\begin{figure*}[!htbp]
    \centering
    \includegraphics[width=\textwidth]{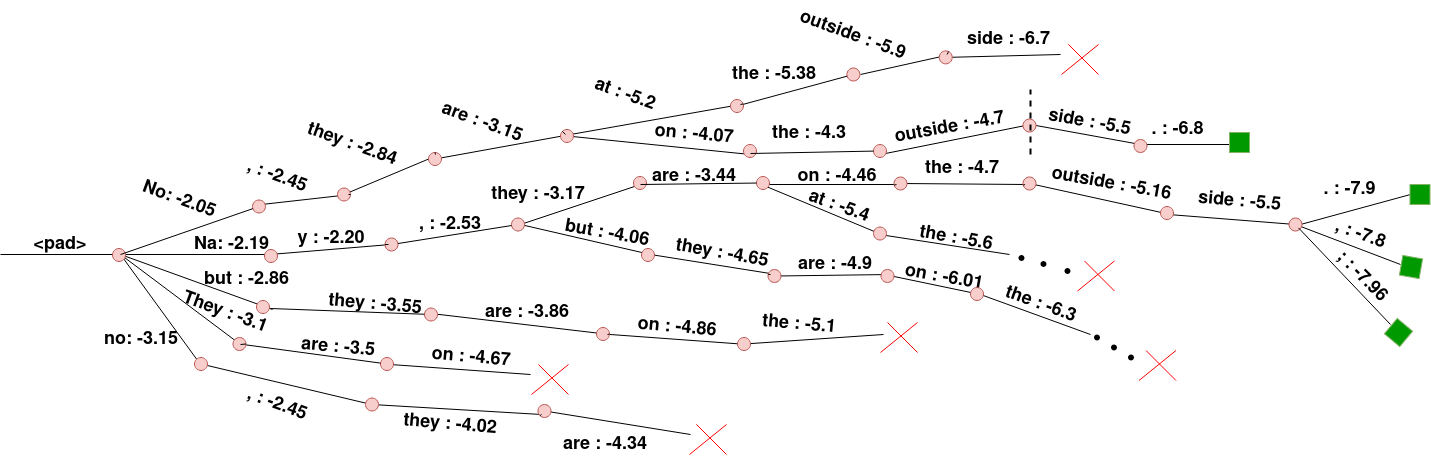}
    \caption{NMT beam search graph of the translation of the German sentence \textit{``Nein, sie sind an der Aussenseite"}. Each arc denotes a hypothesised token and its associated log-likelihood score. The green boxes denote the \textit{B}-best paths with highest overall likelihood scores while the crosses denote the paths that were rejected due to insufficient scores.}
    \label{beamsearch}
\end{figure*}

Beam search is the standard  decoding method used by most auto-regressive language decoding and text generation systems, including NMT. Such systems are based on the  assumption that the probability distribution of a word sequence can be decomposed into the product of conditional next word distributions:
\begin{equation}
     P(w_{1:T} | W_0) = \prod_{t = 1}^{T} P(w_t | w_{1:t-1},W_0) 
\end{equation}
where $W_0$ is the initial context. The length $T$ of the word sequence is generally determined on-the-fly. Beam search works by keeping the most likely  hypotheses at each time step and eventually chooses the hypothesis that has the highest overall likelihood. This addresses the limitations of greedy decoding techniques that select the token with the highest probability at each time step and hence may end up missing high probability words hidden behind a low probability word.

Figure \ref{beamsearch} illustrates a beam search run for MT of a DE sentence, with beam width $B = 5$. Following the initial ``$<$pad$>$" token, the search keeps track of the \textit{B} subsequent most likely paths 
(``$<$pad$>$",``Na"), (``$<$pad$>$",``but"), (``$<$pad$>$",``They"), (``$<$pad$>$", ``no"), (``$<$pad$>$", ``No"). The  algorithm then branches the top-most path in the graph into two paths, and finds that the path (``$<$pad$>$", ``No", ``,", ``they", ``are", ``on") has a higher overall likelihood than the path (``$<$pad$>$", ``No", ``,", ``they", ``are", ``at"). 
The search concludes in B-best paths with highest overall score. The dotted arrow denotes two candidate paths, namely (``$<$pad$>$", ``No", ``,", ``they", ``are", ``on", ``the", ``outside") and (``$<$pad$>$", ``No", ``,", ``they", ``are", ``on", ``the", ``outside", ``side", ``.").

Beam search is, however limited to exploring the search space in a greedy left-right fashion retaining only the \textit{B}-best candidates. This can often lead to repetitive generation of similar tokens across multiple MT  hypotheses. As a result, the \textit{B}-best candidates produced by beam search often differ only slightly from each other, thus failing to capture the inherent ambiguity of real-world language generation tasks.

\section{Diverse Beam Search}

\begin{figure*}[!htbp]
        \centering
        \includegraphics[width=\linewidth, trim={0 0 9cm 0},clip]{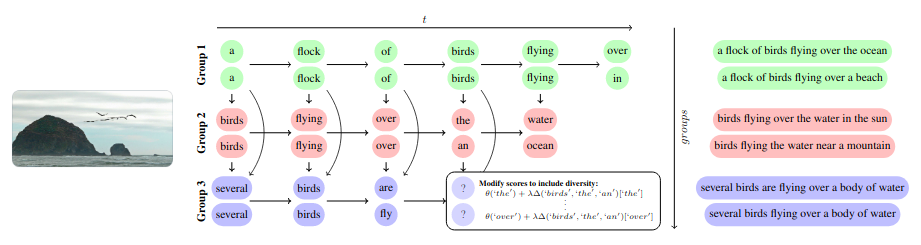}
        \caption{Figure taken from \cite{vijayakumar2016diverse} illustrating  the working of group beam search by operating left-to-right through time and top-to-bottom through groups.}
        \label{dbs}
\end{figure*}
Diverse Beam Search (DBS) \citep{vijayakumar2016diverse} addresses the aforementioned limitations of traditional beam search by decoding a list of diverse outputs and optimizing for a diversity-augmented objective while introducing minimal computational or memory overhead. Formally, DBS works by partitioning the set of all B beams at time \textit{t} into G non-empty disjoint subsets $Y^g_t$ [$g \in G$] such that each partition contains $B' = B / G$ groups. Unlike beam search, DBS  proceeds with the introduction of a dissimilarity term that measures how different a sequence $y_t$ is against a group $Y^g_t$. The search then optimizes each beam group conditioned on the previous ones. Figure \ref{dbs} shows a snapshot of DBS generating a caption for a picture, keeping  $B=6$ and $G=3$. In particular, group 3 is chosen to be propagated forward by computing the diversity-augmented score of all words in the dictionary conditioned on groups 1 and 2.

\section{Related Work}
This section reviews previous works that have targeted learning language models on outputs of machine translation in domain-specific scenarios in particular. Several of these have leveraged best translation as well as N-best translations for training domain-specific LMs on generic MT data \citep{unanue2020pretrained, domhan2017using}. \cite{punjabi2019language} describe the limitations of building LMs for domain-specific tasks in the face of formal-styled web corpora and code switching. The authors circumvent training LM on raw translations by post-editing the translations through resampling of named entities,  applying   domain adaptation through in-domain data selection, and rescoring with in-domain LMs. A similar line of work has also been along improving the performance of MT itself through domain adaptation. \cite{wang2017sentence} propose such a domain adaptation technique for NMT by exploiting an RNN's internal embedding of the source sentence and using the embedding's similarity to select the sentences closer to in-domain data.  

\begin{figure*}[!htbp]
    \centering
    \begin{subfigure}[t]{0.5\textwidth}
        \centering
        \includegraphics[width=2.5in, keepaspectratio=true]{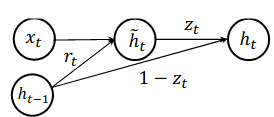}
        \caption{Standard GRU architecture}
        \label{person_corr}
    \end{subfigure}%
    ~ 
    \begin{subfigure}[t]{0.5\textwidth}
        \centering
        \includegraphics[width=2.5in, keepaspectratio=true]{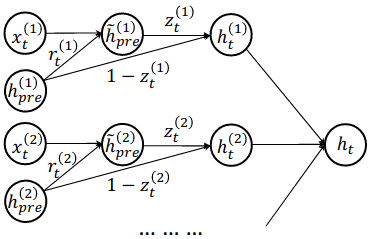}
        \caption{Word-lattice based GRU architecture}
        \label{activity_corr}
    \end{subfigure}
\caption{Figure taken from \cite{su2017lattice} illustrating the differences between a normal GRU and GRU for word-lattices.}
\label{grus}
\end{figure*}

Unlike the existing works, this thesis checks on the advantages of using an N-best list of translations over the 1-best translation. We hypothesize that an MT decoded beam search graph could be better than a mere N-best sequence of translations as it contains scores of alternate  competing words. On similar lines, prior works have attempted to train n-gram LMs on graphs decoded by ASR. \cite{kuznetsov2016learning}, for instance, compare the differences between using full lattices and 1-best ASR output for computing the probability of occurrence of an n-gram in an uncertain data sample. Previous works have also proposed RNNs for graph inputs. \cite{su2017lattice} proposed a word-lattice based RNN encoder for NMT that alleviates the impact of tokenization errors for languages without natural word delimiters (e.g., Chinese). Figure \ref{grus} shows their modified Gated Recurrent Unit (GRU), a type of RNN unit, that can compactly encode multiple alternative hypotheses $x^{(k)}_t$ from an ASR lattice in order to generate a pooled RNN hidden state $h_t$. 

Similarly, \cite{sperber2017neural} propose LatticeLSTMs which leverage lattice posterior scores building upon TreeLSTM's \citep{tai2015improved} child-sum and forget gates. Moreover, \cite{jagfeld2017encoding} presented a GRU-based dialogue state tracking system which operates on ASR confusion networks. More recently, \cite{pal2020modeling} discuss the benefits of using confusion networks over N-best ASR list for neural dialogue state tracking by encoding the 2-dimensional confusion network into a 1-dimensional sequence of embeddings using attentional confusion network encoders.

\chapter{Language Models for Machine Translated Data}
\label{ch:approach}
\section{Problem Statement}
Our end goal lies in training task-specific LMs for ASR. In the absence of training data for the target language of interest, we rely on task-specific text from another language and resort to machine translation to obtain task-specific training data in the target language. This gives rise to a two-fold problem: (a) MT on domain-specific text can be less accurate since we use a pre-trained MT model that has been trained on generic parallel corpora, and (b) an LM trained on texts translated through an MT model might not provide a good coverage of different ways of formulating spoken queries in the target language.

\begin{figure*}[h!]
    \centering
    \begin{subfigure}[t]{\textwidth}
        \centering
        \includegraphics[width=0.6\textwidth]{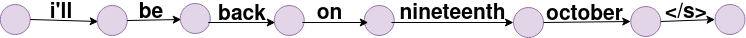}
        \caption{1-best transcription sequence}
        \label{firstsausage}
    \end{subfigure} \\
    ~ 
    \begin{subfigure}[t]{\textwidth}
        \centering
        \includegraphics[width=\textwidth]{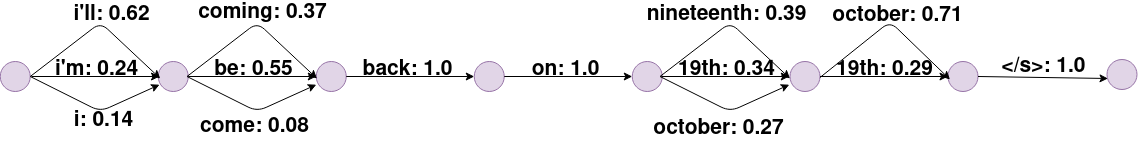}
        \caption{Confusion network
        derived form machine translation with 3-best beam search paths.
        }
        \label{secondsausage}
    \end{subfigure}
\caption{Graph representation of the 1-best translation and the confusion network built from N-best translations.}
\label{problemstatement}
\end{figure*}
Moving beyond the 1-best translations of an MT system, N-best translations can help us tackle the problems of inaccurate MT and the lack of diversity in the MT outputs. Furthermore, we would like to exploit the scores of the multiple competing paths in an MT decoder. We achieve this by deriving a word confusion network from the N-best translations or the MT decoder's beam search graph. Figure \ref{problemstatement} presents such a sample 1-best translation and a corresponding word confusion network. As evident in the figure, a confusion network consists of a sequence of  bins associated with alternative word hypotheses and their associated posterior probability scores.

\begin{figure*}[h!]
        \centering
        \includegraphics[width=0.9\textwidth]{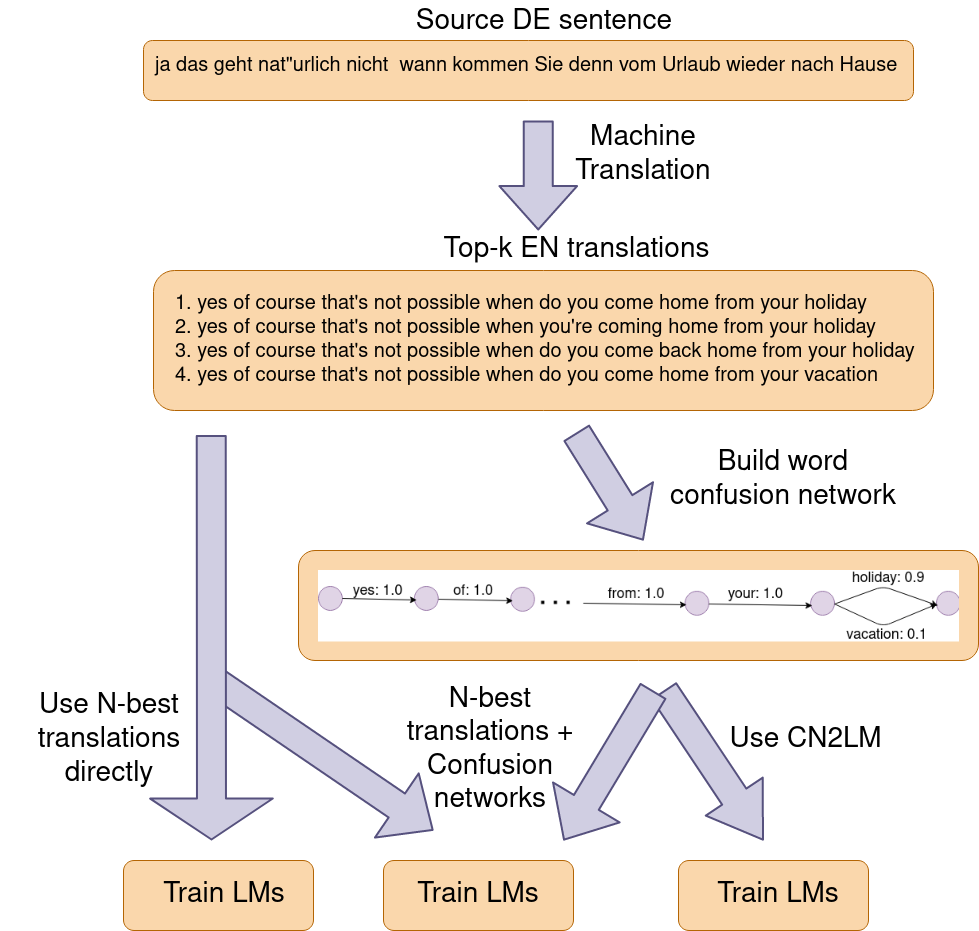}
        \caption{
        Schema of our approach to train task specific LMs using N-best translations, a confusion network, and a combination of both.}
        \label{blockdia}
\end{figure*}

Finally, Figure \ref{blockdia} presents a schema of our approach to train task specific LMs in a target language with the help of machine translation. 

\section{Training Language Models on Confusion Networks}
\label{sec:cn2lm}
In this section, we describe the method adopted for training n-gram and RNN LMs on confusion networks derived from beam search graphs of an MT decoder. We employ the publicly available CN2LM library\footnote{\url{https://gitlab.inria.fr/comprise/speech-to-text-weakly-supervised-learning}} to perform this training. The CN2LM library has been developed under the COMPRISE project\footnote{\url{https://www.compriseh2020.eu}} with the aim of training language models from ASR confusion networks decoded from speech data. The CN2LM package provides features for training  3-gram LMs as well as RNN LMs. 

\subsection{n-gram Language Models on Confusion Networks}
\label{ngramlmcn2lm}
Let \textbf{u}\textit{w} denote the n-gram word sequences occurring in a text corpus where u denotes the context window comprising of the $n-1$ previous words for the word \textit{w}. Then n-gram LM probabilities can be estimated as:
\begin{equation}
    P(w|\textbf{u}) = \frac{c(\textbf{u}w)}{\sum_{w} c(\textbf{u}w)}
\end{equation}
where $c(.)$ denotes the n-gram counts. However, obtaining practically usable LMs requires the application of smoothing methods over naive count-based estimates. Among the best performing smoothing methods for n-gram LMs has been the modified interpolated Kneser-Ney (KN) smoothing \citep{chen1999empirical}. A hindrance at applying KN smoothing to posterior scores of arcs is that the scores are fractional (see Figure \ref{secondsausage}). To address this, \cite{zhang2014kneser} proposed a modified interpolated KN smoothing (ieKN) which was then used by \cite{levit2018expect} to learn n-gram LMs from crowdsourced and ASR transcriptions. 

The CN2LM library extends the ieKN approach to ASR confusion networks.  It extracts n-gram bin sequences from the confusion networks and populates different possible word sequences of length \textit{n}. A score is then assigned to each n-gram sequence by multiplying the associated arc posteriors and finally applying: (a) the ieKN smoothing to obtain the \textit{n}-th order probability estimates, and (b) a recursive smoothing technique to obtain the lower-order probability estimates.

\subsection{RNN Language Models on Confusion Networks}
\label{rnnlmcn2lm}
If a corpus $W$ is to be denoted as a word sequence $x_1, x_2, x_3, ..., x_t, ...$ then a word-based RNN LM with parameters $\theta = \{U, V, W\}$ can be modelled mathematically as:
\begin{equation}
    h_t = \delta(W h_{t-1} + U x_t)
\end{equation}
where $h_t$ and $h_{t-1}$ denote the RNN hidden states modeling the context observed until time step tokens $x_t$ and $x_{t-1}$, $U$ denotes the weight matrix for input to hidden connections and $W$ denotes the weight matrix for hidden-to-hidden recurrent connections. The outputs of the network are given as:
\begin{equation}
\label{softmax}
        p(\hat{x}_{t+1} | h_t) = \text{softmax}(V h_t)
\end{equation}
where $V$ denotes the weight matrix parameterizing the hidden-to-output connections $\delta(.)$ is a non-linear activation function, and the softmax function calculates the word-level LM probabilities $p(\hat{x}_{t+1}|h_t)$. The training objective is given as:
\begin{equation}
\label{loss}
    \hat{\Theta} = \argminA_\theta  \sum_{h} \sum_{j} - \log p(\hat{x}_{t+1}^{j} | h_t)
\end{equation}

where \textit{j} loops over the length of the sentence. However, conventional RNN LMs cannot be applied to 
confusion networks given that each time step $t$ in a confusion network corresponds to a confusion bin with distinct alternative word hypotheses (see Figure \ref{secondsausage}). Equation \eqref{softmax}  thus needs to be adapted so as to first compute a hidden state $h_t^i$ for each possible arc $w^i_t$ in a bin followed by a pooling of states:
\begin{align}
    h^i_t &= \sigma(W h_{t - 1} + V x^i_t) \\
    h_t &= \text{pool}_i(h^i_t)
\end{align}
where the plausible choices for $\text{pool}()$ include mean, max or attention-based mechanisms. Subsequently, the LM loss function needs to handle multiple possible output arcs $x^j_{t+1}$ for the time step $t+1$. In CN2LM, this is addressed by computing the Kullback-Leibler (KL) divergence between the RNN predictions and the confusion bin posteriors $p(x^j_{t+1}|o_t)$. This can be formulated as:
\begin{equation}
        \hat{\theta} = \argminA_\theta  \sum_{t} \underset{j }{\text{KL}} (p(x^j_{t + 1}| o_{t+1}), p(\hat{x}^j_{t+1} | h_t))
\end{equation}
where $j$ is the index over the LM vocabulary and $o_{t+1}$ is the observed speech signal leading to the confusion bin posteriors.

\section{Word Confusion Networks from Beam Search}
\label{sec:beamsrch2cn}
One of the main contributions of this thesis is a method to derive ASR-like word confusion networks from a beam search decoding graph of an MT decoder. We describe the procedure to derive such confusion networks in the following steps.
\begin{enumerate}
    \item Computation of N-best lists and the log-likelihoods: N-best list and the log-likelihood for each of the N-best hypotheses are obtained from the beam search graph. The log-likelihoods are computed by summing the log-likelihoods of each token in the corresponding beam search path.
    \item Post-processing of N-best lists: We observe that the N-best lists might contain inconsistencies owing to the limitations of the beam search decoding process. The steps listed below describe how we alleviate these:
     \begin{enumerate}
         \item Stripping off punctuation: All but those punctuation marks present inside a token itself, e.g. ``it's", are retained.
         \item Pruning paths with end-of-sentence tokens: Everything after the first appearance of  the end-of-sentence token ``$<$/s$>$" or ``$<$pad$>$" token in a path is removed, including the token itself.
         \item Pruning unusually longer paths: Beam search graphs occasionally
         contain repetitive sequence of tokens, which lead to unusually longer outputs. We prune such paths automatically to remove the repeated sequence of tokens.
         Formally, if $L_S$ and $L_T$ are the lengths of the source sentence ``$S$" and translated sentence ``$T$", then $T$ is said to be of unusual length if $L_S - L_T \geq \delta$ where $\delta$ is a parameter dependent upon how much the length of ``$T$" might vary relative to ``$S$". In our experiments on DE-EN translations, we safely assume $\delta = \text{max}(3, 1 + L_S / 5)$.\footnote{This builds on our general observation that EN translation of a DE sentence is no longer than 20\% of its length.} If a repetitive sub-sequence is then found in $T$, we strip if off limiting the maximum length of $T$ to $L_S + \delta$.
     \end{enumerate}
    \item N-best list to confusion network:
    We use SRILM toolkit's \textit{nbest-lattice} tool\footnote{\url{http://www.speech.sri.com/projects/srilm/manpages/nbest-lattice.1.html}} to derive a confusion network from the N-best lists and the log-likelihood scores for each path.
     \item Handling *DELETE* tokens: \textit{n-best lattice} introduces \textit{*DELETE*} tokens for arcs that are suspected to be missing in a given candidate translation but is present in other translations. We first remove all such bins which do not contain arcs other than those labelled with *DELETE* tokens. Moreover, the scores of all the remaining arcs labelled with *DELETE* at a given time step are summed up and reassigned so that the score for all *DELETE* tokens at a given step remains identical.

    \item Post-processing of confusion networks
    \begin{enumerate}
       
        \item Handling digits, cases and Out of Vocabulary (OOV) tokens: All digits are marked as $<$d$>$ and all tokens which are OOV are marked as $<$unk$>$. If multiple $<$d$>$ or $<$unk$>$ arcs remain in a confusion bin then their they are collapsed into a single arc and the scores for these arcs are summed together.
        
          \begin{figure*}[h!]
        \centering
        \includegraphics[width=0.9\textwidth]{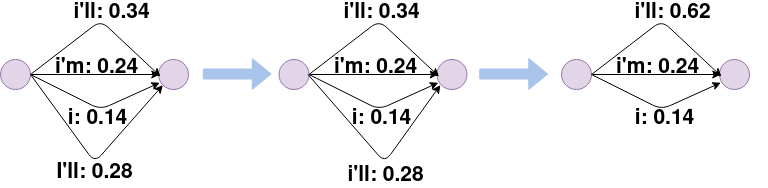}
        \caption{A snapshot of postprocessing cases in an individual arc of confusion network.}
        \label{postprocess_sausage}
\end{figure*}

    Figure \ref{postprocess_sausage} shows such an instance of a postprocessing step wherein the cases are handled first followed by collapsing of the arcs labelled with ``i'll" into one.
    
    \item Normalization of bin scores: The last step of involves  re-normalization of the scores in each confusion bin such that the scores in each bin sum to 1.0.
    \item Sorting of arcs: Following normalization, the bins at each time step are arranged in descending order of the arc scores.
    \end{enumerate}
\end{enumerate}


\chapter{Experiments and Evaluation}
\label{ch:exp}

In order to verify our hypothesis, that N-best translations and their scores can help to train better task specific LMs in a target language, we train and evaluate n-gram and RNN LMs on an standard English-German (EN-DE) chat translation dataset. 
We consider a scenario in which task specific German chats are available and the goal is to build English LMs for the same task. 
CN2LM based n-gram and RNN LMs, as described in Section \ref{sec:cn2lm}, are compared with baseline n-gram and RNN LMs trained over N-best sequences of MT outputs. Moreover, we also compare translations obtained from normal beam search and group beam search, discussed in Section \ref{beamsearchgraphs}.

\section{Experiment Setup}
\subsection{Dataset}
For all our experiments, we use the English-German (EN-DE) chat translation corpus from the shared task held at the Fifth Conference on Machine Translation (WMT20)\footnote{\url{http://www.statmt.org/wmt20/chat-task.html}}. Based on the work of \cite{byrne2019taskmaster}, the corpus includes English and German chats across six domains: ordering pizza, movie tickets and coffee drinks, making auto repair appointments and restaurant reservations, and setting up ride services. The corpus has a standard train, dev, and test set split with the statistics as mentioned in Table \ref{tab:datasets}.

\begin{table}[!htbp]
    \centering
    \caption{Statistics for the WMT20 dataset.}
    \begin{tabular}{ |c|c| } 
     \hline
     \# Train sentences & 13222 \\ \hline
     \# Dev sentences & 1822 \\ \hline
     \# Test sentences & 2017 \\ \hline
     \# Avg. sentence length & 9 \\ \hline
     \# \% OOV & 19\% \\ \hline
     \end{tabular}
    \label{tab:datasets}
\end{table}

The OOV words denote the percentage of  dev set words that are not present in the train set. As a pre-processing step prior to feeding the sentences to an MT model, we remove all the punctuation from the source train sentences except for ``:", ``," and ``." which may occur in the context of time (``8:30 am"), prices (``\$ 5,40"), and acronyms (``p.m.").

\subsection{MT setup}
For MT, we experimented with two open-source state-of-the-art EN-DE models, from 
OpusMT \citep{TiedemannThottingal:EAMT2020} and OpenNMT \citep{klein-etal-2017-opennmt}. We find that the OpusMT model  achieves better translation quality (BLEU score of 33.52) on our DE-to-EN translation task, as compared to the OpenNMT model (BLEU score of 21.48). Therefore, we proceed with OpusMT as our choice of NMT model. Translation outputs, both on the train as well as the dev set, undergo the same post-processing as discussed in Section \ref{sec:beamsrch2cn}, \textit{i.e.}, we remove all the punctuation present outside a word, replace the OOV tokens with $<$unk$>$ and replace digits with $<$d$>$. 

\subsection{Evaluation}
We evaluate the trained language models on the post-processed dev set using perplexity  as the choice of our evaluation metric. Although an intrinsic evaluation method, perplexity is standard measure used to compare LMs for ASR.  Perplexity on an evaluation
set $W = {w_1, w_2, ..., w_N}$ can be defined as the inverse probability of observing the evaluation set normalized by the number of words in the set. Mathematically,
\begin{equation}
    \text{Perplexity}(W) = \frac{1}{P(w_1, w_2, ..., w_N)^\frac{1}{N}}
\end{equation}
As a result, minimizing perplexity of an evaluation set is equivalent to maximizing the probability of observing the set according to the LM. 

\subsection{Model and Training Configurations}
\label{config_model}
All our RNN LMs use a single layer of Gated Recurrent Units (GRUs). Input and output word embedding matrices are shared and tied during training. We set the dimensional of word embeddings to be 64. Similarly, the hidden states of the GRU are set to 64. Training uses a batch size of 32. We employ the Adam optimizer with an initial learning rate of 0.001. The learning rate is scheduled to reduce by a factor of 0.1 if the loss does not decrease for 10 epochs. We leverage an early stopping strategy, on the dev set perplexity, with a patience set to 15 epochs. The maximum number of arcs in a confusion bin is restricted to 5 and the confusion bin scores are re-normalized to sum to 1.0.

\subsection{Implementation Notes}
All our experiments were performed on a machine with Intel Xeon E5-2650 v4 CPU with 12 cores, 128 GB RAM, and two Nvidia GTX 1080 Ti graphics card. We use PyTorch 1.6.0 to train all RNN LMs. OpusMT build is based upon the Hugging Face Transformers \citep{wolf2019huggingface} implementations. Group beam search implementation from Hugging Face Transformers was used. We rely on the publicly available SRI Language Modeling toolkit to train classical n-gram LMs.\footnote{\url{http://www.speech.sri.com/projects/srilm/}} 

\section{Results}
The perplexity scores for translations on normal beam search and group beam search are shown in Table \ref{nbs_results} and Table \ref{dbs_results}, respectively. From both the tables, we observe a common trend of decreasing perplexity as the number of N-best translations increases from 1 to 100. When comparing the n-gram LMs, we see that CN2LM n-gram offers lower perplexity compared to training n-gram LMs on N-best translations, both for normal beam search and group beam search. 
\begin{table}[!htbp]
    \centering
    \caption{Perplexity of LMs trained on N-best translations from normal beam search.}
    \begin{tabular}{ |c|| p{1.5cm}|p{1.5cm} || p{1.5cm}|p{1.5cm}|p{1.75cm}| } 
    \hline
	\multirow{2}*{N (in N-best)} & \multicolumn{2}{p{2cm}||}{n-gram LM} & \multicolumn{3}{p{3cm}|}{RNN LM} \\ \cline{2-6}
     & N-best & CN2LM & N-best & CN2LM & CN2LM + N-best \\ \hline
     1 & 56.5 & 56.2 & 66.5 & 66.5 & 62.8 \\
     5 & 50.1 & 48.4 & 57.8 & 63.8 & 57.9 \\
     10 & 49.2 & 45.6 & 56.3 & 61.64 & 52.9 \\
     20 & 47.2 & 43.8 & 51.7 & 59.2 & 50.5 \\
     50 & 46.6 & 43.0 & 50.1 & 57.5 & 47.9 \\
     100 & 43.7 & 42.9 & 48.0 & 56.9 & 47.1 \\
    \hline    
    \end{tabular}
    \label{nbs_results}
\end{table}

\begin{table}[!htbp]
    \centering
    \caption{Perplexity of LMs trained on N-best translations from group beam search.}
    \begin{tabular}{ |c|| p{1.5cm}|p{1.5cm} || p{1.5cm}|p{1.5cm}|p{1.75cm}| } 
    \hline
	\multirow{2}*{N (in N-best)} & \multicolumn{2}{p{2cm}||}{n-gram LM} & \multicolumn{3}{p{6cm}|}{RNN LM} \\ \cline{2-6}
     & N-best & CN2LM & N-best & CN2LM & CN2LM + N-best \\ \hline
     1 & 55.5 & 55.2 & 64.3 & 65.8 & 63.3 \\
     5 & 55.5 & 55.2 & 64.3 & 65.8 & 63.3 \\
     10 & 53.3 & 52.1 & 61.8 & 63.2 & 59.7 \\
     20 & 50.3 & 47.5 & 56.9 & 64.5 & 55.5 \\
     50 & 47.8 & 43.0 & 51.3 & 57.3 & 49.9 \\
     100 & 45.9 & 43.0 & 49.9 & 57.3 & 48.4 \\
    \hline    
    \end{tabular}
    \label{dbs_results}
\end{table}

When comparing the RNN LMs, we see that CN2LM RNN results in higher perplexity scores compared to RNN LMs trained on N-best translations. However, supplementing CN2LM with the N-best translations  leads to a major boost in CN2LM training and results into perplexity scores lower than the RNN LMs trained on N-best translations. Comparison of perplexities resulting from normal beam search versus group beam search shows that N-best translations from normal beam search result in lower perplexities. This is counter-intuitive to the hypothesis that a diverse beam search resulting into diverse translations should result into better LMs.

We note that RNN-based LM lags behind n-gram based LM on N-best lists as well as on confusion networks across both the beam search. This can be attributed to the fact that we avoid hyperparameter tuning of the RNN models and instead stick to a single set of hyperparameters  offering  a decent performance (see Section \ref{config_model}). As a result, we compare the performance of n-gram LMs independent of the RNN LMs.

\section{Discussion}
These results show that the use of confusion networks with competing word hypotheses can help achieve lower perplexity scores for LMs than using 1-best and N-best MT output sequences. Our findings concerning the role of N-best translations and their posterior scores build upon the work of \cite{sperber2017neural}. They used  LatticeLSTM as an encoder unit in an attentional encoder-decoder model to achieve consistent improvements with speech translation lattices over baselines comprising either 1-best hypothesis or the word lattice without posterior scores. 

The lower perplexity scores for LMs trained on confusion networks show that the presence of posterior scores can help mitigating the confusion arising from MT errors. The approach we describe for obtaining word confusion networks takes advantage of the pre-existing MT beam search graphs. An immediate benefit of this strategy remains the integration of such a module into standard MT systems with minimal overhead. We show that exploiting the MT beam search graphs further offers the flexibility of obtaining such confusion network across other variants of beam search such as the diverse beam search.

As domain-specific ASR systems keep gaining popularity, there is a growing need to train LMs that  provide better coverage of queries while being as flexible as generic LMs for deployment. Although identifying the gains from word confusion networks represents a crucial step towards bridging the gap due to limited domain specific data in our language of interest, much work needs to be done in order to achieve practical use cases. For instance, while we make use of N-best translations and diverse beam search, our framework still remains limited by the diversity of source sentences themselves. A suitable next step for our work thus lies in verifying that the input to MT systems are rich enough.

\chapter{Conclusion}
\label{ch:conclusion}

This thesis evaluated an approach for learning task specific LMs for automatic speech recognition in scenarios where task specific training data is not available in target language but in another language. The approach mainly relies on the use of pre-trained machine translation. We showed that using N-best translations of the task specific data result in LMs with lower perplexities as compared to using only the 1-best translation. We further developed a procedure to derive MT word confusion networks, consisting of alternative and competing word hypothesis along with their scores, from N-best translations and their corresponding log-likelihoods. As part of evaluation, n-gram and RNN LMs were trained on these MT confusion networks using the CN2LM library developed under the COMPRISE project. We found that CN2LM n-gram LMs trained on MT confusion networks performed better than n-gram LMs trained on N-best translations. Moreover, supplementing CN2LM training with N-best translations results in RNN LMs with lowest perplexities.

As a future direction to this research, we plan to evaluate the LMs trained on MT confusion networks in a task-specific ASR setting, in order to demonstrate their effectiveness to reduce ASR word error rates. We further plan to experiment with data augmentation strategies such as backtranslation to induce diversity among the original source language transcriptions prior to feeding these to an NMT system.

\section{Acknowledgement}
This work was supported by the European Union's Horizon 2020 Research and Innovation Program under Grant Agreement No. 825081 COMPRISE (\url{https://www.compriseh2020.eu/}). Experiments were carried out using the Grid'5000 testbed, supported by a scientific interest group hosted by Inria and including CNRS, RENATER and several Universities as well as other organizations (see \url{https://www.grid5000.fr}).

\addcontentsline{toc}{chapter}{Bibliography}
\bibliographystyle{plainnat}
\bibliography{thesis}


\end{document}